\documentclass[runningheads]{llncs}
\usepackage{graphicx}

\usepackage{tikz}
\usepackage{comment}
\usepackage{amsmath,amssymb} 
\usepackage{color}

\begin{document}
\pagestyle{headings}
\mainmatter
\def\ECCVSubNumber{5}  

\title{RPT: Learning Point Set Representation for Siamese Visual Tracking} 

\author{Ziang Ma \and
Linyuan Wang \and
Haitao Zhang \and
Wei Lu \and
Jun Yin
}

\authorrunning{Z. Ma et al.}
%
\institute{Zhejiang Dahua Technology, Hangzhou, China\\
\email{\{ma\_ziang,wang\_linyuan,zhang\_haitao1,lu\_wei,yin\_jun\}@dahuatech.com}}
\maketitle

\begin{abstract}
While remarkable progress has been made in robust visual tracking, accurate target state estimation still remains a highly challenging problem. In this paper, we argue that this issue is closely related to the prevalent bounding box representation, which provides only a coarse spatial extent of object. Thus an efficient visual tracking framework is proposed to accurately estimate the target state with a finer representation as a set of representative points. The point set is trained to indicate the semantically and geometrically significant positions of target region, enabling more fine-grained localization and modeling of object appearance. We further propose a multi-level aggregation strategy to obtain detailed structure information by fusing hierarchical convolution layers. Extensive experiments on several challenging benchmarks including OTB2015, VOT2018, VOT2019 and GOT-10k demonstrate that our method achieves new state-of-the-art performance while running at over 20 FPS. 
\end{abstract}

\section{Introduction}

Robust visual tracking is a fundamental task for various computer vision
applications, including video monitoring analysis, man-machine
interaction and intelligent driving. It aims at locating an arbitrary
target of interest during a whole video sequence. Although substantial
progress has been made in recent years, the design of a high-performance
tracker is still highly challenging due to moving camera, occlusions and
variations in structure.

Much attention has been invested for accurately estimating the target
state in recent years. Its difficulty lies in frequently changed
appearance caused by target or camera movement and varying postures. A
region proposal network (RPN) was introduced to estimate the target
bounding box with a pre-defined set of anchor boxes~\cite{siamrpn,siamrpn++,dasiamrpn}. However, it hinders
the generalization and efficiency of the Siamese-based tracking
framework. The no-prior box design was subsequently presented~\cite{siamfc++,siamban,siamcar}, which is free of prior
knowledge about target scale/ratio disrtibution. It
directly views locations as training samples and predicts the relative
offsets to the corners or sides of bounding box. The location is
assigned with a positive label if it falls within a preset center area
of the ground-truth bounding box, ignoring the target appearance and
geometric structure. Besides, the target state is commonly described
with a bounding box for the conveniences of feature extraction and
ground truth annotation. It provides only a coarse spatial
extent of object, and lacks the modeling capability for geometric
transformations, thereby severely restricting the localization accuracy~\cite{reppoints,densereppoints}.

\begin{figure}[t]
\centering
\includegraphics[width=\textwidth]{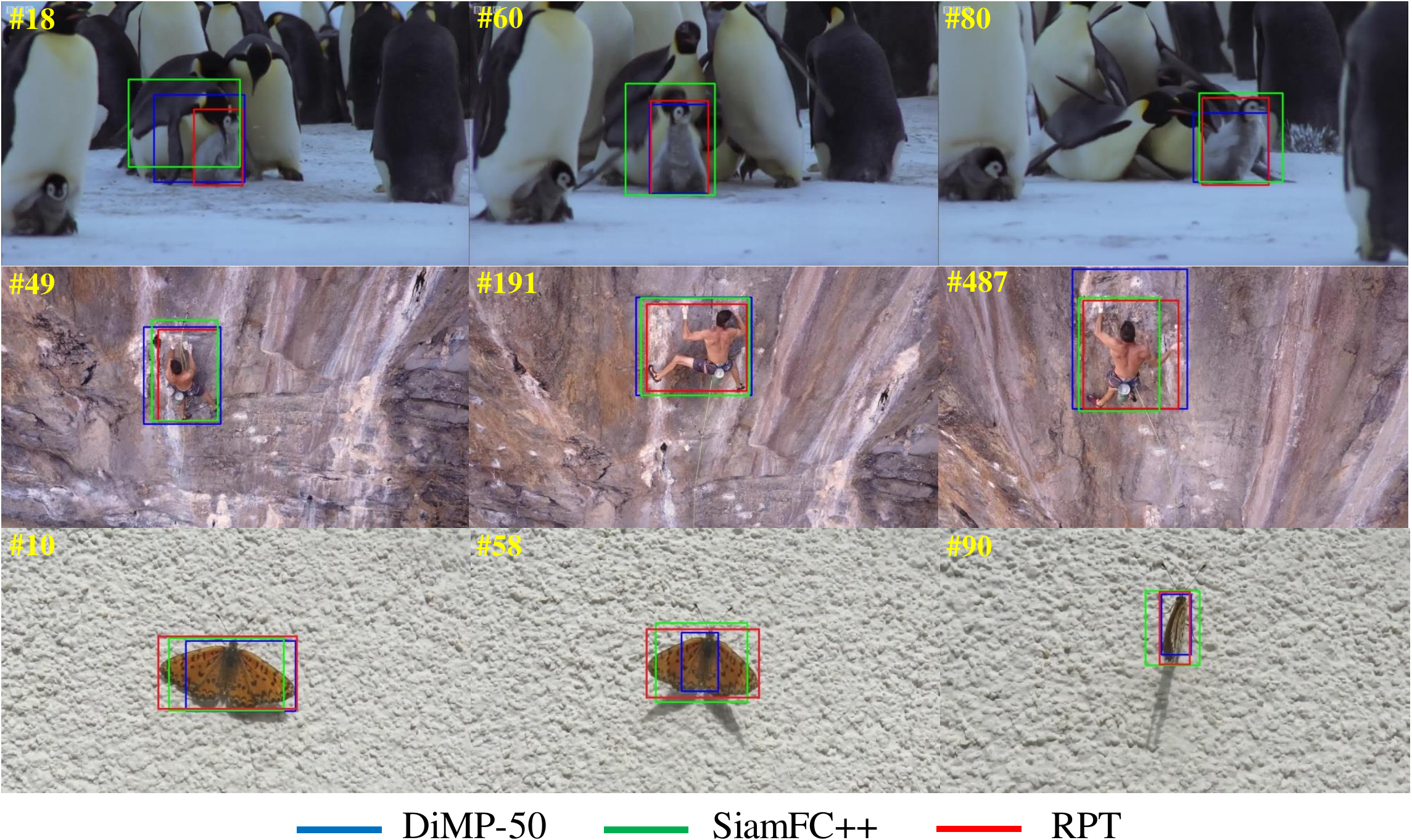}
\caption{Qualitative comparisons with state-of-the-art trackers. Our approach RPT obtains more accurate bounding box predictions when handling variations in posture, scale and aspect ratio}
\label{fig:trackingresults}
\end{figure}

In this paper, we propose a novel tracking method named Representative
Points based Tracker (RPT) to address the issue of accurate target
estimation. RPT models the target state with a new finer
representation as a set of representative points, and learns to arrange
them to indicate the target's spatial extent and geometrically
significant positions. In contrast to the coarse encoding of bounding
box, the point set representation facilitates more fine-grained
localization and modeling of object appearance. The RPT framework
is constructed with two parallel branches, one primarily
accounting for target estimation with the point set representation, the
other trained online to provide high robustness against distractors.

\begin{figure}[t]
\centering
\includegraphics[width=\textwidth]{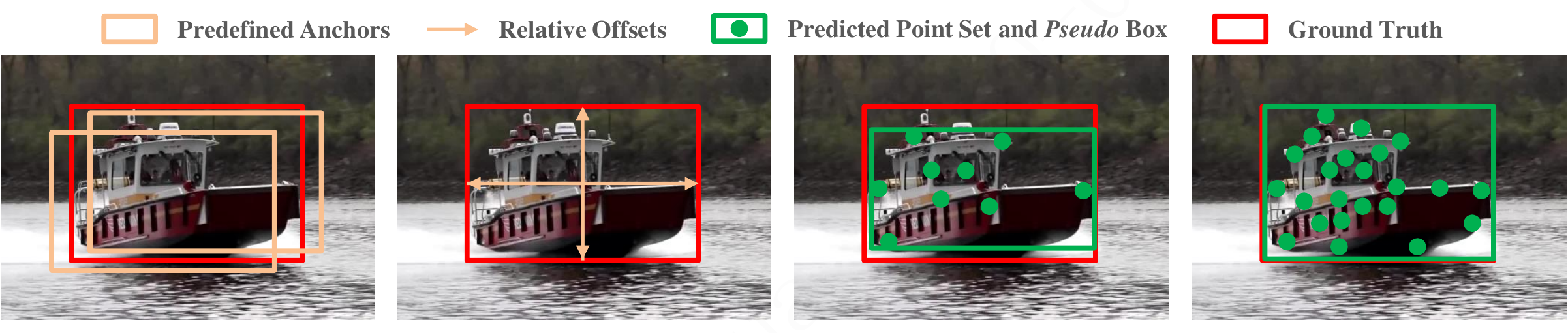}
\caption{Comparison for methods used to estimate the target state. Multi-level aggregation provides detailed structure information of objects, facilitating more fine-grained localization}
\label{fig:target_estimate}
\end{figure}

The main contributions of this paper are threefold.

$\bullet$ A novel point set representation facilitating more accurate target
  estimation is employed in the field of visual tracking.

$\bullet$ We aggregate hierarchical convolution layers to provide detailed
  structural information of target and high discriminative
  power when handling distractors.

$\bullet$ Our work achieves new state-of-the-art performances and runs at 20 FPS
  on several benchmarks, including OTB2015~\cite{OTB2015}, VOT2018~\cite{VOT2018}, VOT2019~\cite{VOT2019} and
  GOT-10k~\cite{GOT-10k}.

\section{Related Work}
Visual tracking is a fundamental computer vision problem, which can be divided into a foreground-background classification task and a target state estimation task. The former case is responsible for distinguishing the target appearance from distractors and the surrounding background. The latter one aims to accurately describe the target region against variations in posture, scale and aspect ratio.

Discriminative correlation filters~\cite{srdcf,csrdcf,drt} and online learning approaches~\cite{atom,drol} have remarkably advanced the target classification task in recent years. On the other hand, accurately estimating the target state is still challenging and severely limited with a multi-scale search strategy~\cite{kcf,eco,siamfc,drt}. The target state is usually represented with a bounding box, and further estimated with various methods. SAMF\cite{li2014a} adopts a multiple scales searching strategy to address the issue of scale variations. GOTURN\cite{goturn} designs a box regression strategy with the Laplace distribution hypothesis, which is restricted to small motions and limited scale changes. SiamRPN and the succeeding works~\cite{siamrpn,siamrpn++,dasiamrpn} introduce a region proposal network (RPN) to regress the target region from pre-defined anchors. It leads to an apparent degeneration in tracking efficiency and generalization. ATOM~\cite{atom} iteratively refines the coarse initial location to obtain a higher overlap between the predicted bounding box and ground truth. Inspired from anchor-free detection pipeline, a no-prior box design is utilized to predict offsets from candidate locations to the corners or sides of desired bounding box~\cite{siamfc++,siamban,siamcar}.

In order to describe more spatial details of objects, rotated bounding box and segmentation mask are further utilized to represent the target state in visual tracking~\cite{ldes,siamMask,D3S}. LDES simultaneously estimates the orientation and scale variation of target via polar coordinate transformation~\cite{ldes}. SiamMask produces a binary segmentation mask via an extra branch trained on YouTube-VOS\cite{youtubevos}, which is a large video dataset with pixel-wise annotations~\cite{siamMask}. D3S obtains a noticeable advancement in segmentation accuracy by constructing a parallel structure with complementary geometric models~\cite{D3S}.

\section{RPT Framework}

\begin{figure}[t]
\centering
\includegraphics[height=6.5cm,width=\textwidth]{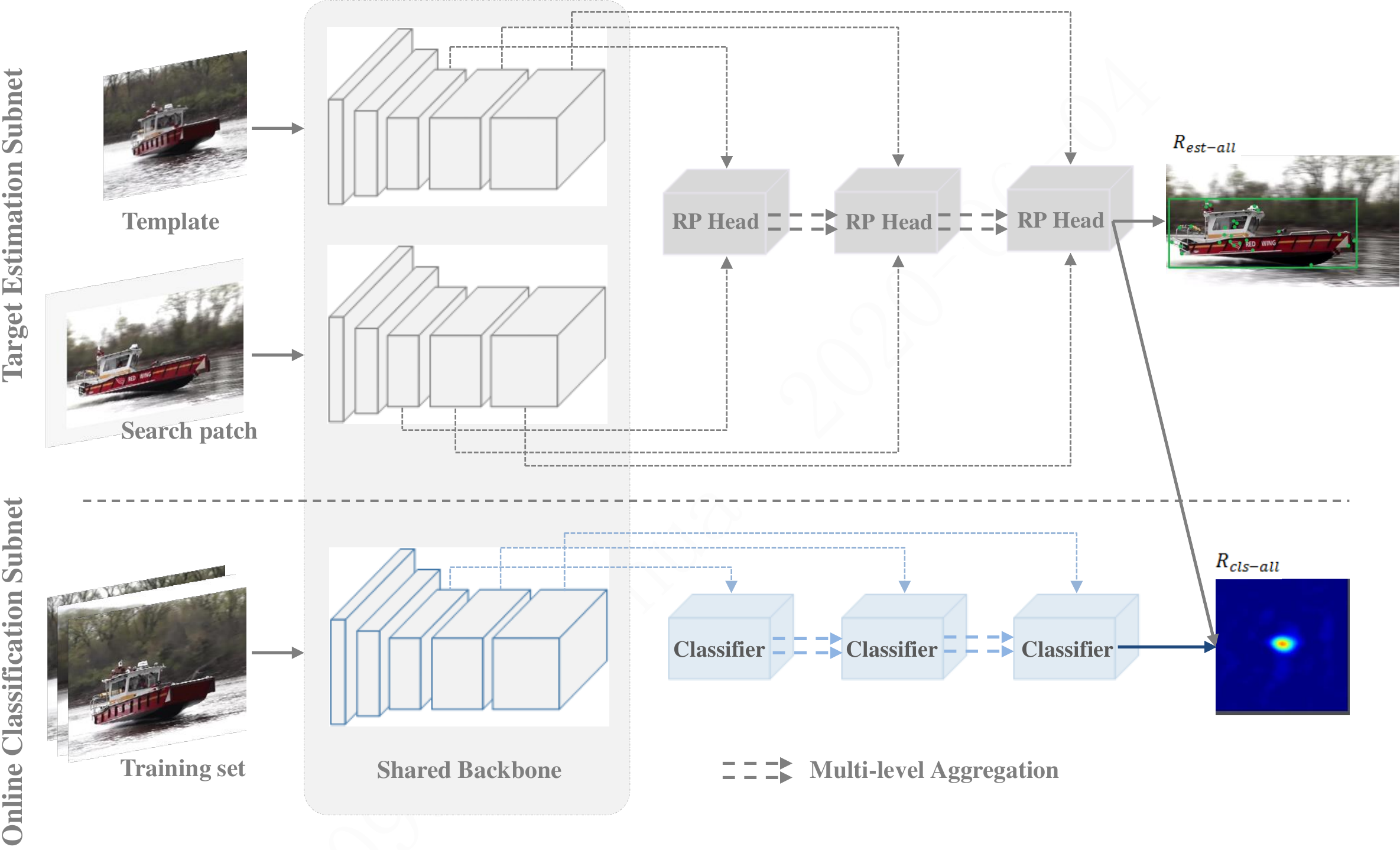}
\caption{The RPT framework with multi-level aggregation. The target estimation subnet predicts the spatial extent of object with a set of representative points in an offline-trained embedding space. The online classification subnet is responsible for distinguishing the target appearance from distractors and the surrounding background}
\label{fig:RPT_structure}
\end{figure}

The RPT framework is constructed with a shared backbone network for
feature extraction, and two parallel subnets accounting for
target estimation and online classification respectively, as illustrated in Figure~\ref{fig:RPT_structure}. Following the architectural
design guidelines in~\cite{CIR}, we adopt ResNet-50~\cite{resnet50} as the backbone
network, and extract hierarchical convolutional features from the last
three residual blocks for multi-level prediction. The target estimation
subnet is driven by a finer representation of object as a point set,
which provides more fine-gained localization. The online classification
subnet is trained exclusively online to enhance the discriminative capability in the presence of distractors.

\subsection{Target Estimation with Point Set Representation}

\begin{figure}[t]
\centering
\includegraphics[width=\textwidth]{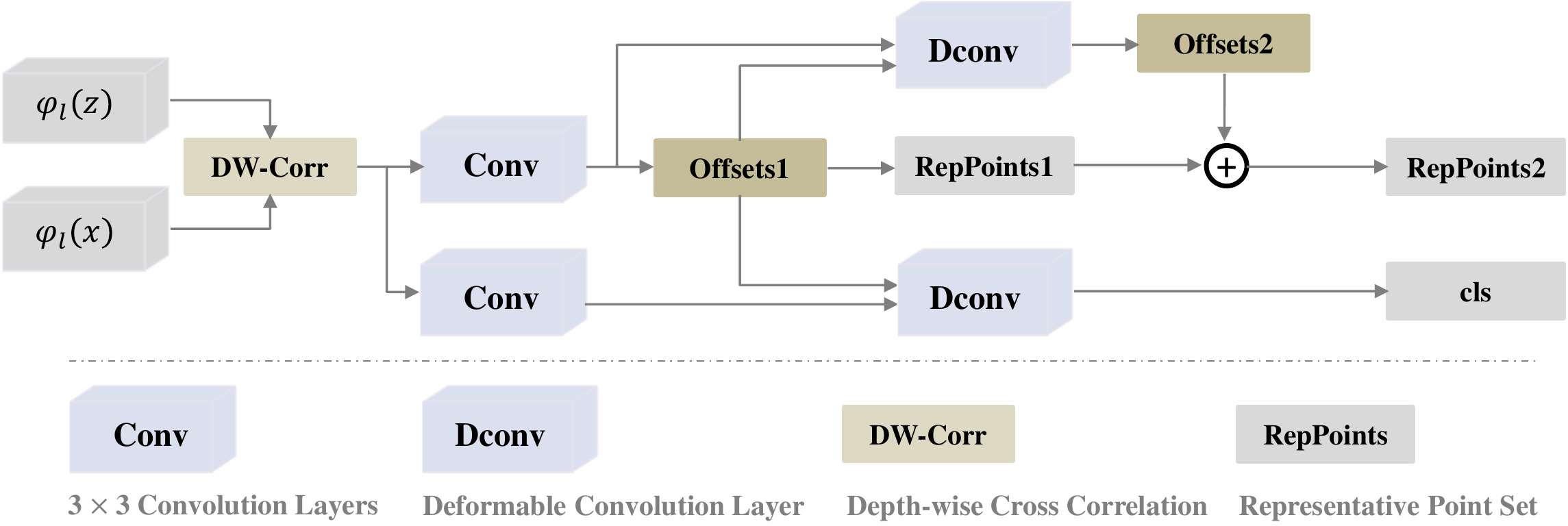}
\caption{The architecture of RP Head}
\label{fig:RPHead}
\end{figure}

In contrast to object detection, tracking target can be an arbitrary object with unknown class. We therefore exploit Siamese-based feature extraction and matching for the requirement of target-specific estimation. Multi-level features from a target template (denoted as z)
and a search region (denoted as x) are extracted. For each feature level, a correlation map between the target patch and the search patch
is obtained via depth-wise cross correlation~\cite{siamrpn++} as:

\begin{align}
g_{l}\left( z,x \right) = \varphi_{l}\left( z \right)*\varphi_{l}\left( x \right)
\end{align}
where $\varphi_{l}\left( z \right)$ and $\varphi_{l}\left( x \right)$
represent the corresponding feature outputs of the \emph{l}-th level.

Inspired from~\cite{reppoints}, we design a classification head and a two-stage
regression head over the correlation map. The head architecture is illustrated in Figure~\ref{fig:RPHead}.
Each location of the correlation map is regarded as a target candidate. For the location $\left(i,j\right)$, the target state is modeled with a set of sample points, which are
uniformly initialized with the corresponding position as:

\begin{align}
R = \left\{ \left( x_{k},y_{k} \right) \right\}_{k = 1}^{n}
\end{align}
\begin{align}
x_{k} = i, y_{k} = j, k = 1,2,...,n
\end{align}
where \emph{n} represents the capacity of the point set. For each candidate, the regression head outputs a set of offsets to refine the
distribution of sample points, while the classification head
outputs two-channels for foreground-background classification. Specifically, the initial sample points are progressively refined
with extra offsets in a point-wise manner:

\begin{align}
R_{r} = \left\{ \left( x_{k} +\Delta x_{k},y_{k} +\Delta y_{k} \right) \right\}_{k = 1}^{n}
\end{align}
where the predicted offsets
$\left\{ \left( \Delta {x}_{k},\Delta {y}_{k} \right) \right\}_{k = 1}^{n}$
are paired with the deformable convolution~\cite{Dconv}. In contrast to
its standard counterpart, deformable convolution augments the regular
spatial sampling locations with additional offsets. It has the
capability of modeling various geometric transformations for scale,
aspect ratio and rotation. The kernel size of deformable convolutions is commonly set as $3\times3$. The number of representative points \emph{n} is accordingly set to 9 in this work.

The refinement process is offline-trained with a multi-task loss for
simultaneous localization and recognition. In the field of visual
tracking, the target region is usually annotated with a bounding box,
which is inconsistent with the point set representation. In order to
utilize the bounding box annotations for supervision, we perform a
min-max operation over the refined point set obtaining a pseudo
box as:

\begin{align}
B_{p} = \left( \min\left\{ x_{k} +\Delta x_{k} \right\},\min\left\{ y_{k} +\Delta y_{k} \right\},\max\left\{ x_{k} +\Delta x_{k} \right\},\max\left\{ y_{k} +\Delta y_{k} \right\} \right)
\end{align}

In our work, the ratio of the overlapping area between the induced
pseudo box and ground-truth bounding box to the union area is utilized
to represent the regression loss. The focal loss~\cite{focalloss} focusing on hard
examples is further employed for foreground-background classification.
Driven by both target regression and classification losses, a set of
representative points are automatically learned that indicates the
object boundaries and semantically prominent regions.

\subsection{Discriminative Online Classification}
The classification head discussed above is employed to distinguish foreground from background. However, it lacks the capacity to
discriminate the target object from similar surrounding instances. In this section, we propose to complement the target estimation pipeline
 with an online-trained classifier to provide high robustness against
distractors. Similar to~\cite{atom,drol}, it is modeled as a light weight
2-layer fully convolutional neural network for efficiency.

The online classifier is trained with a certain amount of target regions
obtained from the last few frames, which share the same size to the
search patch. The training sample is commonly labeled with
a two-dimensional Gaussion function centered at the target position, where
the geometric structure of object is neglected. Here, we propose to
label the confidence of target presence according to the representative
point set. It is offline-learned to indicate the spatial extent of
object and semantic key points. The desired classification score for
each pixel of the training sample is therefore constructed by
calculating the average position deviation to the point set.

After obtaining the online classification score, we combine it with the
classification head output of the target estimation subnet as

\begin{align}
f_{l} = \alpha f_{\textit{online}}^{l} + \left( 1 - \alpha \right)f_{\textit{offline}}^{l}
\end{align}
where\(\ f_{*}^{l}\) represents the corresponding response maps for the
\emph{l}-th level, and \(\alpha\) is the parameter controlling the
impact of these two confidence scores.
\subsection{Multi-level Aggregation}
As pointed by~\cite{HCFST,HCFT}, convolutional features extracted from earlier
layers preserve finer spatial information benefiting for precise
localization, while latter activations capture rich semantic information
facilitating robustness against variation in appearance. Thus
we propose to employ hierarchical convolutional layers contributing to the inference of both online classification and target estimation.

For online classification, the per-pixel confidences of target
presence obtained from each classifier are combined via a
weighted-fusion layer as

\begin{align}
R_{cls - all} = \sum_{l = 3}^{5}{w_{l}*f_{l}}
\end{align}
The outputs of the last three residual blocks share the same
spatial resolution, weighted sum is therefore implemented in a pixel-wise manner.
The set of weights \(w_{l}\) are end-to-end optimized
together with the network.

For the target estimation subnet, each head outputs only a small set of
representative points, which are insufficient when handling complicated
object structures. Tracking accuracy is also limited as an inaccurate
pseudo box is obtained from the sparse point set. Thus we propose
to utilize a significantly larger set of points as the target state
representation. The dense point set is simply constructed as a
collection of representative points obtained by each head:

\begin{align}
R_{est - all} = \bigcup_{l = 3}^{5}R_{l},\ R_{l} = \left\{ \left( x_{k}^{l},y_{k}^{l} \right) \right\}_{k = 1}^{n}
\end{align}
As shown in Figure~\ref{fig:target_estimate}, the fusion of sample points provides more
elaborate structure of objects that is beneficial for accurate target
estimation.

\section{Experiments Results}

\subsection{Implementation Details}

Following the architectural design guidelines in~\cite{CIR}, ResNet-50~\cite{resnet50}
pre-trained on ImageNet~\cite{imagenet} is adopted as our backbone network.
Multi-level features from the last three residual blocks are extracted
for aggregation. The down-sampling operations are removed in the
\emph{conv4} and \emph{conv5} blocks to preserve finer spatial
information. Meanwhile, dilated convolutions with different
atrous rates are exploited to improve the receptive fields.

The target estimation subnet is trained offline with image pairs
selecting from YouTube-BoundingBox~\cite{youtubebb}, COCO~\cite{coco} and ImageNet
VID~\cite{imagenet}. For the first stage regression, location $\left(i,j\right)$ on the correlation map is considered as a positive sample
if its corresponding location on the search region is closest to the center of tracking target.
The second stage regression and
classification are then conducted on the first set of representative points. If the IOU (intersection-over-union)
between the induced pseudo box and ground-truth is larger than 0.5, it is assigned with a positive label, and
if the IOU is smaller than 0.4, it is assigned with a negative label, and otherwise ignored. For both stages,
only the positive samples are exploited for the target state regression.

The training process is driven by SGD on 8 GPUs with 128
image pairs per mini-batch for 20 epochs. We employ a warm-up learning
rate linearly ascended from 0.001 to 0.005 for the first 5 epochs, while
the learning rate for the last 15 epochs is exponentially decayed from
0.005 to 0.0005. In the first 10 epochs, only the head architecture of
target estimation is trained. The whole network is fine-tuned in the
last 10 epoch, where the learning rate of backbone is set to be 10 times
smaller than the base value. The online classification subnet is trained
with target regions obtained from the last few frames. Conjugate
Gradient~\cite{atom} is employed as the optimizer for efficient online learning. Our
tracker is implemented on a PC with a GeForce GTX 1080Ti GPU using
PyTorch.

\subsection{Comparison with SOTA}
The proposed RPT is compared with state-of-the-art trackers on several
popular tracking Benchmarks, including OTB2015~\cite{OTB2015}, VOT2018~\cite{VOT2018}, VOT2019~\cite{VOT2019} and
GOT-10k~\cite{GOT-10k}.

\begin{figure}[t]
\centering
\includegraphics[width=0.45\linewidth]{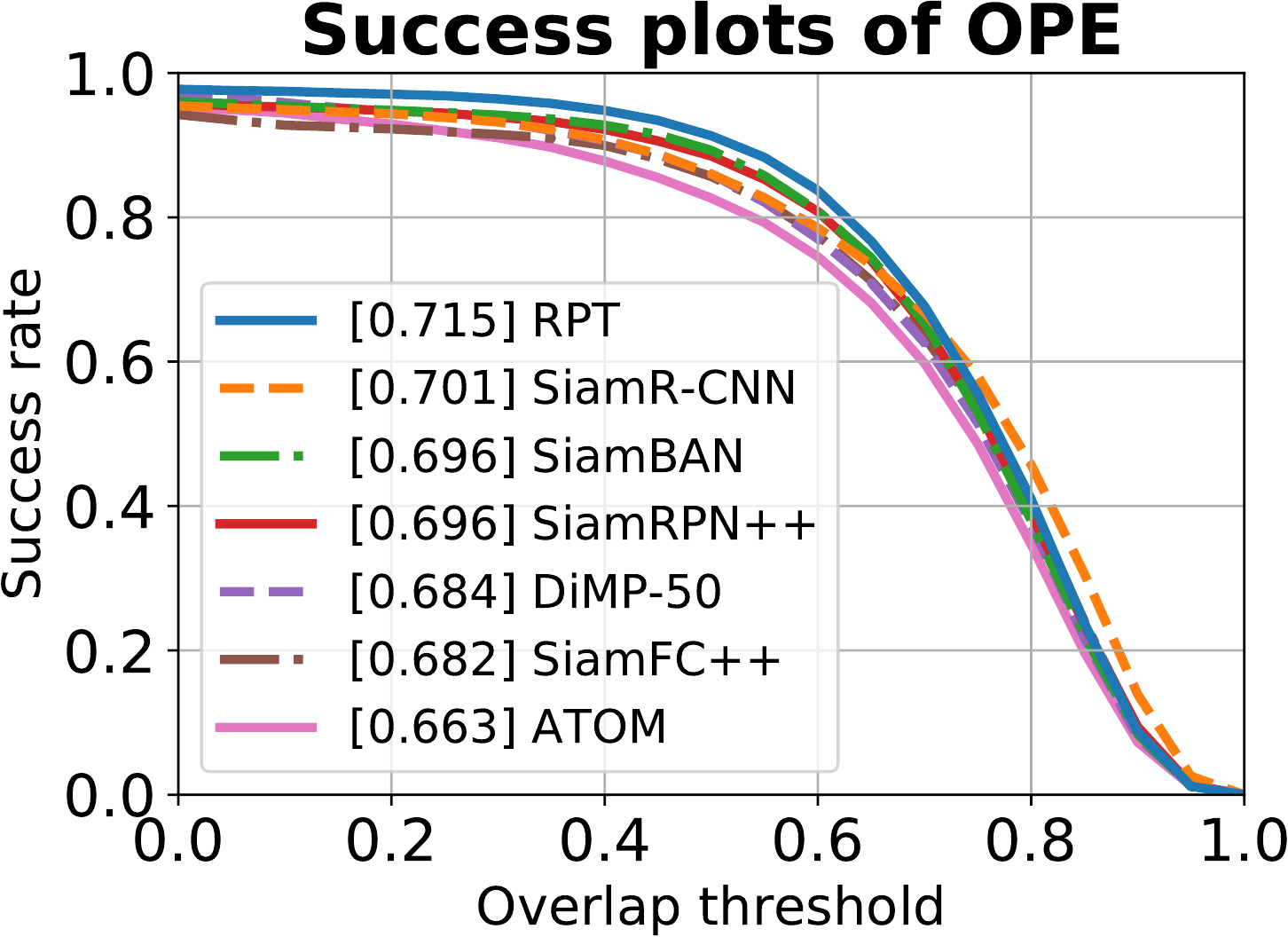}
\includegraphics[width=0.45\linewidth]{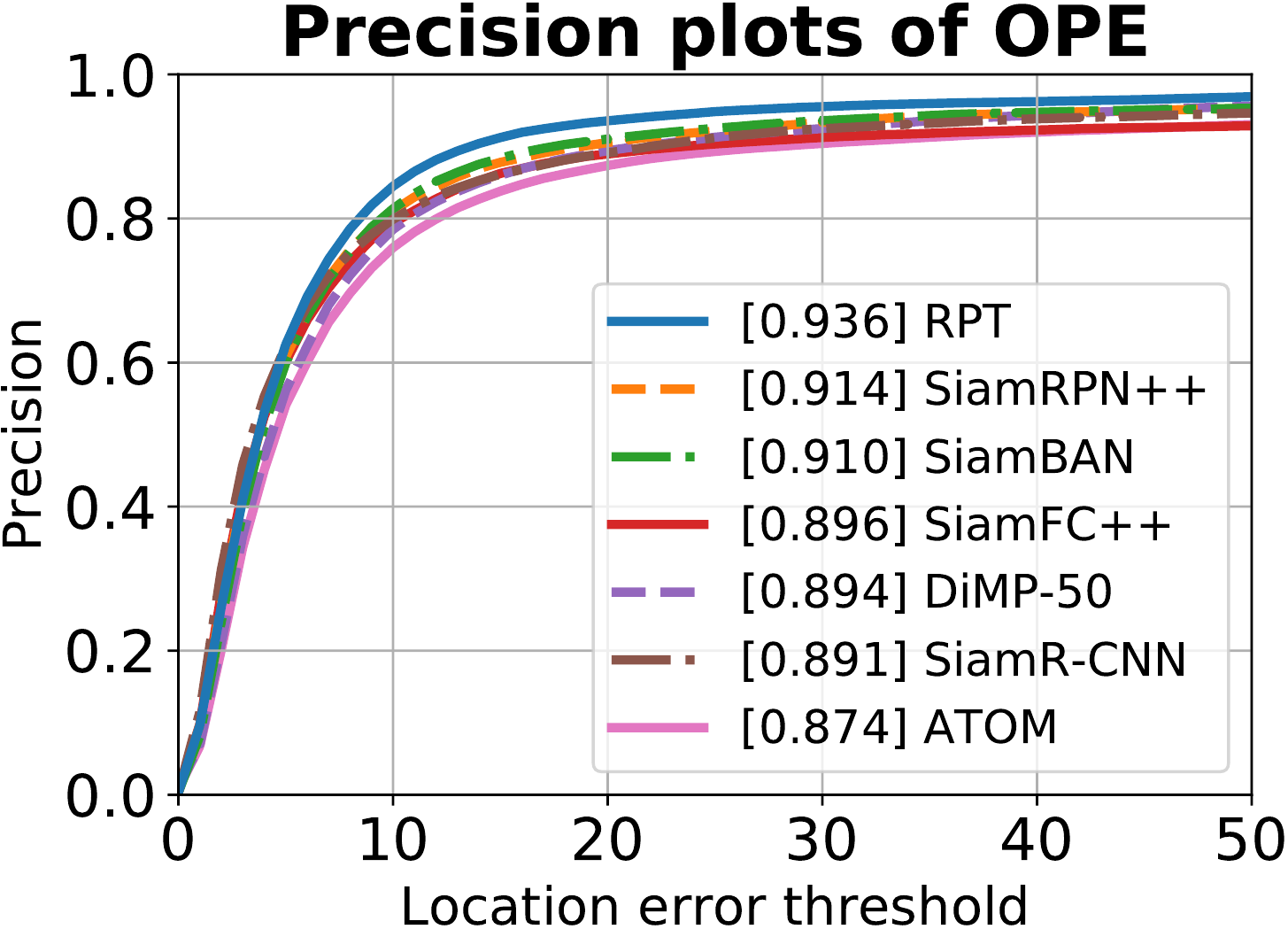}
\caption{Success and precision plots on OTB2015}
\label{fig:OTBresults}
\end{figure}

\textbf{OTB2015.} OTB2015~\cite{OTB2015} is one of the most widely used datasets for
quantitative evaluation in visual tracking. We compare RPT with numerous
recent and advanced visual trackers, including SiamRPN++~\cite{siamrpn++}, ATOM~\cite{atom}, DiMP~\cite{dimp},
SiamFC++~\cite{siamfc++}, SiamR-CNN~\cite{siamrcnn} and SiamBAN~\cite{siamban}, in terms of precision score and success
rate. The precision score is formulated as the fraction of frames where
the distance from the center of tracking result to the ground truth is
under 20 pixels, while the success rate denoted as the proportion of
frames whose intersection-over-union with the ground truth surpasses
0.5. As illustrated in Figure~\ref{fig:OTBresults}, our tracker obtains an AUC (area under
curve of success plot) score of 0.715 and precision score of 0.936,
which are 1.4 and 2.2 percentage points higher than the previous best
result of SiamR-CNN~\cite{siamrcnn} and SiamRPN++~\cite{siamrpn++}.

\textbf{GOT-10k.} GOT-10k~\cite{GOT-10k} is a large-scale dataset for tracking
evaluation, which contains almost over 10 thousand sequences in training
subset and 180 sequences in both validation and test subsets. Average
overlap (AO) is employed for evaluations on this benchmark. For fair
comparison, methods are required to be trained using the given dataset.
The proposed RPT achieves a competitive result compared with the
state-of-the-art methods with an AO score of 0.624, as shown in Table~\ref{table:got-10k}.

\begin{table}
\begin{center}
\caption{Comparison with state-of-the-art trackers on GOT-10k. The best two results are marked in \textcolor{red}{red} and \textcolor{blue}{blue} fonts}
\label{table:got-10k}
\begin{tabular}{@{}lccccccc@{}}
\hline\noalign{\smallskip}
& ATOM~\cite{atom} & SiamFC++~\cite{siamfc++}  &D3S~\cite{D3S}& DiMP-50~\cite{dimp} & SiamR-CNN~\cite{siamrcnn} & RPT &\tabularnewline
\noalign{\smallskip}
\hline
$SR_{0.5}$($\uparrow$)& 0.635 & 0.695 & 0.676 &0.717& \textcolor{blue}{0.728} &\textcolor{red}{0.730} &\tabularnewline
$SR_{0.75}$($\uparrow$)& 0.402 & 0.479 & 0.462 &0.492& \textcolor{red}{0.597}& \textcolor{blue}{0.504}&\tabularnewline
AO($\uparrow$)& 0.556 & 0.595 & 0.597&0.611& \textcolor{red}{0.649} & \textcolor{blue}{0.624}&\tabularnewline
\hline\noalign{\smallskip}
\end{tabular}

\end{center}
\end{table}

\textbf{VOT2018}. VOT2018~\cite{VOT2018} consists of 60 sequences with various
challenging factors. In contrast to the aforementioned datasets, it's
annotated with a rotated bounding box and restarted the tracker in case
of failures. Methods are measured with three evaluation metrics, namely
accuracy (A), robustness (R) and expected average overlap (EAO). Table~\ref{table:vot}
shows the comparison with the existing top-performing approaches on
VOT2018~\cite{VOT2018}, in which RPT attains the best robustness and EAO among all
methods. D3S~\cite{D3S} is superior in terms of accuracy as numerous segmentation
sequences from Youtube-VOS~\cite{youtubevos} are utilized to produce a binary mask.

\textbf{VOT2019}. VOT2019~\cite{VOT2019} is refreshed by replacing the 12 least
difficult sequences from the previous version with several carefully
selected sequences in GOT-10k dataset. The same measurements are
exploited for performance evaluation. As illustrated in Table~\ref{table:vot}, we
achieve an accuracy of 0.623, a robustness of 0.186 and an EAO of 0.417,
outperforming DRNet~\cite{VOT2019} (leading performance on the public dataset of
VOT2019 challenge) and other SOTA methods in terms of all metrics.

\begin{table}
\begin{center}
\caption{Comparison with state-of-the-art trackers on VOT2018 and VOT2019. The best two results are marked in \textcolor{red}{red} and \textcolor{blue}{blue} fonts}
\label{table:vot}
\begin{tabular}{@{}l|ccc|ccc@{}}
\hline
&  &VOT2018 & &  &VOT2019 &\tabularnewline
& EAO($\uparrow$) & R($\downarrow$) & A($\uparrow$) & EAO($\uparrow$) & R($\downarrow$) &
A($\uparrow$)\tabularnewline
\hline
ATOM~\cite{atom} & 0.401&0.204 &0.590 &0.292 &0.411&0.603\tabularnewline
SiamR-CNN~\cite{siamrcnn} &0.408 &0.220 &0.609 &- &- &-\tabularnewline
SiamRPN++~\cite{siamrpn++} &0.414 &0.234 &0.600 &0.285 &0.482 &0.599\tabularnewline
SiamFC++~\cite{siamfc++} &0.426 &0.183 &0.587 &- &- &-\tabularnewline
DiMP-50~\cite{dimp} & 0.440&0.153 &0.597 & 0.379&0.278 &0.594\tabularnewline
SiamBAN~\cite{siamban} &0.452 &0.178 &0.597 &0.327 &0.396 &0.602\tabularnewline
SiamAttn~\cite{siamattn} &0.470 &0.160 &\textcolor{blue}{0.630} &- &- &-\tabularnewline
DRNet~\cite{VOT2019}&- &-&-&\textcolor{blue}{0.395}&\textcolor{blue}{0.261}&\textcolor{blue}{0.605}\tabularnewline
D3S~\cite{D3S}&\textcolor{blue}{0.489} &\textcolor{blue}{0.150}&\textcolor{red}{0.640}&-&-&-\tabularnewline
\hline
RPT & \textcolor{red}{0.510}& \textcolor{red}{0.103}&0.629 &\textcolor{red}{0.417} &\textcolor{red}{0.186} &\textcolor{red}{0.623}\tabularnewline
\hline\noalign{\smallskip}
\end{tabular}
\end{center}
\end{table}

\subsection{Ablation Study}\label{header-n50}

In this section, an extensive ablation study is conducted to verify the
impact of individual components on OTB2015~\cite{OTB2015}. The modified version of
SiamFC++~\cite{siamfc++} using ResNet-50~\cite{resnet50} as backbone network is exploited as baseline.
As illustrated in Table~\ref{table:ablation}, the baseline approach obtains an AUC score of
0.675. By gradually adding the components of point set representation,
online classification and multi-level aggregation, we achieve sustained
performance gains of 2.5\%, 0.9\% and 0.6\% in terms of AUC score. It is
proven that the main contributions of our work facilitate more accurate
target estimation. A similar conclusion can be obtained on other
benchmarks.

\begin{table}
\begin{center}
\caption{Ablation Study on OTB2015. Baseline denotes a modified SiamFC++ with ResNet-50 as backbone. PSR, OC and MLA represent the components of point set representation, online classification and multi-level aggregation, respectively}
\label{table:ablation}
\begin{tabular}{@{}cccc|ccc@{}}
\hline
Baseline & PSR & OC & MLA & Precision($\uparrow$) & AUC($\uparrow$) & $\Delta$AUC\tabularnewline
\hline
$\surd$  & $\times$ & $\times$ & $\times$ & 0.898 & 0.675 & - \tabularnewline
$\surd$  & $\surd$  & $\times$ & $\times$ & 0.925 & 0.700 & +2.5\% \tabularnewline
$\surd$  & $\surd$  & $\surd$  & $\times$ & 0.928 & 0.709 & +0.9\% \tabularnewline
$\surd$  & $\surd$  & $\surd$  & $\surd$  & 0.936 & 0.715 & +0.6\% \tabularnewline
\hline\noalign{\smallskip}
\end{tabular}
\end{center}
\end{table}

\subsection{Evaluation on VOT2020 Challenge}\label{header-n52}

In contrast to previous versions, VOT2020 is annotated with segmentation
masks using a new performance evaluation protocol. The novel protocol
avoids tracker-dependent resets and re-defines the VOT basis measures.
For evaluations on segmentation, the RPT framework is complemented with
a modified D3S~\cite{D3S} to obtain class-agnostic object mask. In particular, the
target location channel in D3S~\cite{D3S} is enhanced with the average position
deviations from the representative point set. The augmented version of
RPT achieves an EAO of 0.539 on VOT2020 challenge.



\section{Conclusions}
In this paper, we propose a novel tracking architecture named RPT to accurately estimate the target state with a finer point set representation. The RPT framework consists of two parallel branches, one primarily accounting for accurate target state estimation in an offline-trained embedding space, the other trained online to obtain high discriminative power in the presence of distractors. Besides, multiple convolution layers are aggregated to provide more fine-grained localization and detailed structural information of target. Comprehensive evaluations on various visual tracking datasets indicate that our method outperforms the recent and advanced trackers in terms of robustness and accuracy.


\clearpage
%
%
\bibliographystyle{splncs04}
\bibliography{rpt_submission}
\end{document}